\documentclass[letterpaper, 10 pt, conference]{ieeeconf}  

\IEEEoverridecommandlockouts                              

\usepackage{graphicx}

\usepackage{multirow} 
\usepackage{makecell} 

\usepackage{comment}

\title{\LARGE \bf
Removing Infrastructure Barriers in Human-Robot Collaboration Through Wireless Reconfigurable Cells
}

\author{Emma Takács$^{1,2,*}$, Mátyás Hajós$^{1,2}$, Ádám Juniki$^{1}$, Ádám Fischer$^{1}$, Zoltán Komáromi$^{1}$, \\ Kristóf Abai$^{1}$, Dániel Horváth$^{1,2}$, Sándor Máthé$^{3}$, Konstantinos Kousias$^{4}$, and Bence Tipary$^{1,2}$
\thanks{$^{1}$HUN-REN Institute for Computer Science and Control (HUN-REN SZTAKI), Budapest, Hungary}%
\thanks{$^{2}$Eötvös Loránd University (ELTE), Budapest, Hungary}%
\thanks{$^{3}$Advanced Machines, Tatabánya, Hungary}%
\thanks{$^{4}$University of Oslo, Norway}%
\thanks{$^{*}$Corresponding author: {\tt\small emmatakacs@sztaki.hu}}%
\thanks{A supplementary video demonstrating the workcell is available at \url{https://youtu.be/zobin6oytGk}}%
}
\usepackage{cite}
\usepackage{color,soul}
\usepackage{hyperref}
\usepackage{amsmath}
\usepackage{amsfonts}
\usepackage{tikz}
\newcommand\copyrighttext{%
  \footnotesize \textit{Accepted at the 2026 IEEE/RSJ International Conference on Intelligent Robots and Systems (IROS).} \\
  \copyright~2026 IEEE. Personal use of this material is permitted. Permission from IEEE must be obtained for all other uses, in any current or future media, including reprinting/republishing this material for advertising or promotional purposes, creating new collective works, for resale or redistribution to servers or lists, or reuse of any copyrighted component of this work in other works.
}
\newcommand\copyrightnotice{%
\begin{tikzpicture}[remember picture,overlay]
\node[anchor=south,yshift=10pt] at (current page.south) {\parbox{\textwidth}{\centering \copyrighttext}};
\end{tikzpicture}%
}
\begin{document}
\bstctlcite{IEEEexample:BSTcontrol}

\maketitle
\thispagestyle{empty}
\pagestyle{empty}


\begin{abstract}
\copyrightnotice 
Human-Robot Collaboration (HRC) plays a vital role in dynamic, high mix, low volume industrial scenarios such as remanufacturing, which frequently face workcell rearrangements. Traditional setups are constrained by power and data cabling, restricting modularity and reconfigurations, while the selection of commercial wireless devices suitable for real-time perception and safe collaboration are greatly limited in availability. This paper presents a highly flexible, wireless, 5G-based system that serves as a versatile experimental testbed for applications including remanufacturing, operator training, and user studies. To eliminate infrastructure barriers, the workcell integrates a novel battery-powered, multi-sensor platform prototype prioritizing low-latency transmission and sensor modularity. Additionally, to support operator safety and system adaptability across environmental shifts, the system integrates a computer vision module for object detection and pose estimation, further augmented for robust hand recognition. Trained on synthetic and real data, the model reliably detects oriented grasping poses and human hands across varying lighting and background conditions (with an \mbox{mAP@50--95} of 97.74 $\pm$ 0.10\% and a mean inference time of 12.5 ms). By accurately determining robot target poses and monitoring for human hands, it enhances operator safety while preserving the workcell's high portability across different physical domains. Offloading these computationally intensive tasks to the edge via 5G, the proposed architecture contributes to resolving the bandwidth-latency trade-off. To demonstrate portability, the system was implemented at two internationally distinct sites in Hungary and Norway, and was evaluated across a combination of public and private, as well as Standalone and Non-Standalone 5G infrastructures. The performed network experiments produced results in round-trip response times down to 12 ms in case of compatible network-device pairings, suitable for safe, adaptive HRC. However, these measurements also revealed practical limitations related to interoperability in current 5G deployments that should be addressed in future works. Nonetheless, the achieved results demonstrate that the low latency required for real-time edge control is reachable in these rapidly reconfigurable industrial scenarios. 

\end{abstract}

\section{INTRODUCTION} 
Human-Robot Collaboration (HRC) is increasingly vital in modern industrial environments~\cite{fan2022multi}, especially remanufacturing workflows characterized by high variability. In these settings, combining the cognitive adaptability of human operators with the precision and tirelessness of robots is essential to handle seamless transitions between more deterministic tasks such as assembly, and more stochastic ones such as disassembly and sorting~\cite{sotskov2023assembly}. Despite this necessity, creating adaptable workcells remains a significant challenge as industrial collaborative robot (cobot) arms and their sensor networks typically rely on extensive cabling for power and high-bandwidth data transmissions. This wired physical infrastructure is a significant barrier, highly restricting mobility, limiting the implementation of dynamic layouts, and making rapid workspace reconfiguration both slow and expensive~\cite{chen20216g}.

While standard wireless technologies offer a potential solution to remove these infrastructure barriers, the practical viability of current approaches remains unproven for real-time safety and control~\cite{zanbouri2024comprehensive}. Also, most modern HRC applications demand robust computer vision (CV) and augmented reality (AR) assistants to facilitate adaptive robot path planning and dynamic operator guidance~\cite{alenjareghi2025computer,moya2023augmented}. However, offloading these computationally intensive perception tasks to the edge over conventional wireless technologies often introduces latency spikes that compromise system responsiveness~\cite{ruggeri2021safety,lyu2022impacts}, which results in a latency-bandwidth trade-off.

To overcome these issues, we propose a highly flexible, wireless HRC cell, shown on the left in Fig.~\ref{fig:sp_prototype}, specifically designed to adapt to environmental changes and dynamic workflows. Acting as a comprehensive experimental testbed, the proposed barrier-free cell supports a full spectrum of research and development scenarios, including remanufacturing, assembly and disassembly, operator training, HRC research, and user studies. The novel adaptability of our proposed architecture builds upon the following main contributions:

   \begin{figure}[t]
      \centering
      \includegraphics[width=1.0\linewidth]{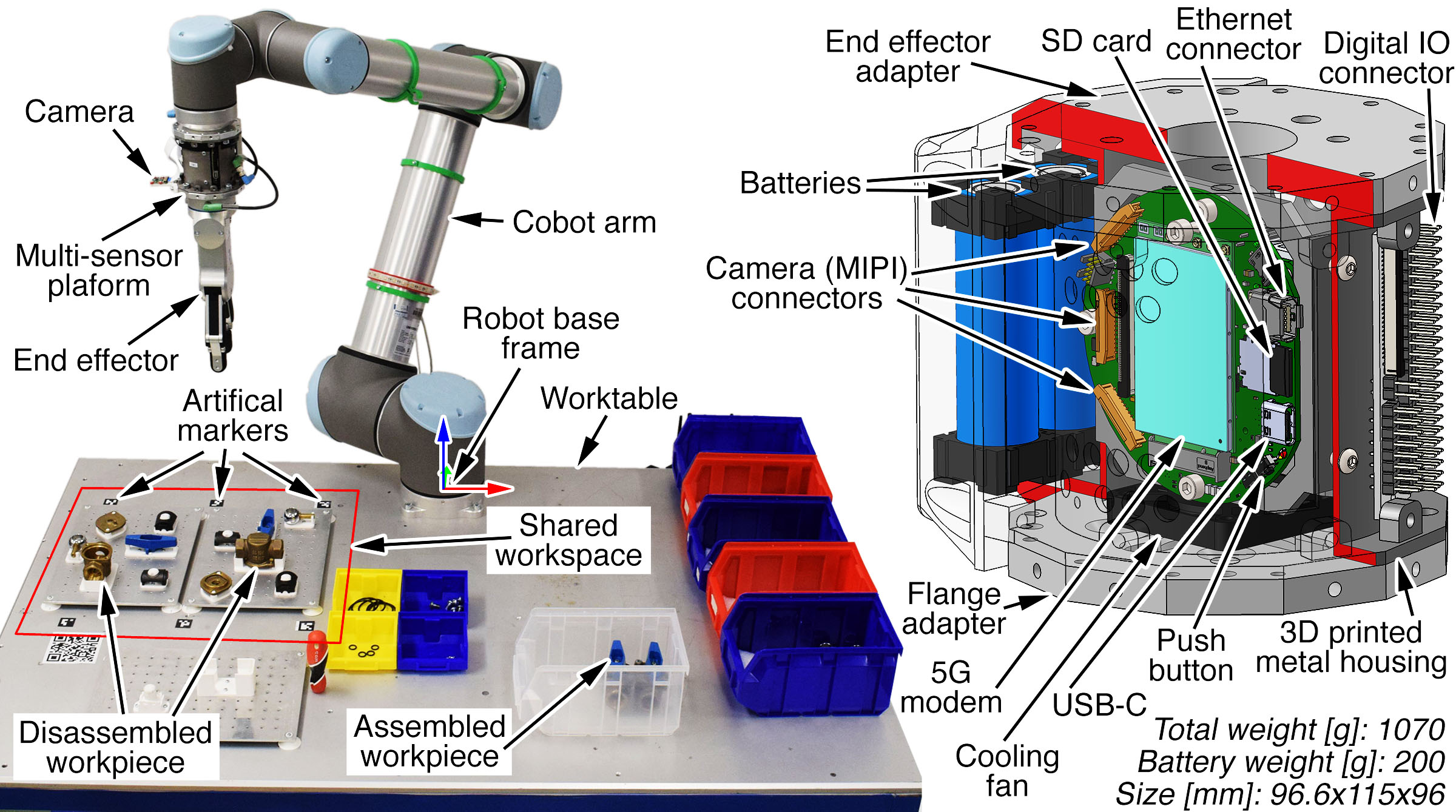}
      \caption{Left: Reconfigurable, wireless, HRC workcell arrangement. Right: 3D model of the designed 5G-enabled multi-sensor platform.}
      \label{fig:sp_prototype}
   \end{figure}
\begin{enumerate}
    \item Robot-mountable multi-sensor platform: We present a battery-operated, cable-free platform prototype designed for rapid reconfiguration. To enhance system modularity, it features connections for up to three cameras and a dedicated IO board for low-speed sensors.
    \item 5G-based wireless communication: The 5G sensor platform offloads computationally intensive CV tasks to the edge without cabling constraints. We empirically evaluated the architecture’s cross-border portability and performance across a diverse set of 5G networks, specifically testing private Standalone (SA), public SA, and private Non-Standalone (NSA) deployments.
    \item Hand-aware, adaptive CV pipeline: The system integrates a perception module for object detection and pose estimation, further augmented for robust hand recognition in HRC scenarios. Trained on synthetic and real data, the model reliably handles varying lighting and background conditions. By determining robot target poses and detecting human hands, it enhances operator safety while maintaining the HRC cell's portability and adaptability across various physical environments.
\end{enumerate}

The rest of this paper is organized as follows. Sections~\ref{sec:problem} and~\ref{sec:related} present the problem statement and related work. In Section~\ref{sec:solution}, our proposed solution approach is outlined, detailing the multi-sensor platform and computer vision design. In Section~\ref{sec:results}, our remanufacturing case study and the corresponding experimental results are shown. Finally, Section~\ref{sec:conclusion} concludes this paper.

\section{PROBLEM STATEMENT} \label{sec:problem}
There are multiple infrastructure barriers in HRC-related applications and shared workspaces. Among these, a major challenge is related to the external wiring of the robot and its sensor networks. The presence of power and data cables can pose safety issues, particularly when human error leads operators to collide or even get entangled in the wiring. 
Furthermore, in accordance with standard safety guidelines for robot integration, exposed cables in shared workspaces are entanglement hazards with nearby objects, which can trigger protective or emergency stops~\cite{iso10218_2}.
This physical limitation is especially problematic in remanufacturing workflows~\cite{xiao2024comprehensive}. Remanufacturing in HRC cells often incorporates both deterministic and stochastic elements originating from its high mix, low volume nature and the handling of components with diverse conditions. The former suggests frequent setup and workspace changes, whereas the latter calls for flexible task execution both on the robot and on the human side. In these highly variable settings, wirings in the HRC system not only increase safety risks but restrict the mobility and adaptability required for efficient remanufacturing~\cite{ruggeri2021safety}. 

Regarding the human operator side, an additional challenge in HRC systems is the absence of reliable human hand-awareness in CV pipelines for shared workspaces. Without accurate detection and tracking of human hands, occlusions and misinterpretations of the workspace may occur, potentially compromising perception accuracy and safe robot behaviour~\cite{fan2022integrated}.

To overcome these physical restrictions for intricate human-machine interactions, transitioning to wireless workcells offers a solution, yet commercially available wireless sensor options are greatly limited. Furthermore, this shift introduces new challenges regarding the heavy computational demands of real-time spatial tracking. Offloading CV tasks to a remote computer over wireless networks often leads to a range of issues including increased latency, bandwidth bottlenecks, and consequently, reduced throughput~\cite{hua2023edge}. 

Moreover, deploying workcells across different areas of a factory or entirely different geographical sites introduces significant environmental variability, such as unpredictable lighting conditions and backgrounds. To ensure that the system can be set up or configured at a single location and seamlessly utilized elsewhere, it requires a robust CV model.

\section{RELATED WORK} \label{sec:related} 
Recent advancements in HRC for complex remanufacturing environments have been heavily driven by researchers such as Xiao et al.~\cite{xiao2024comprehensive,xiao2025intelligent} and Lee et al.~\cite{lee2024review}, who highlight the growing need for real-time, data-intensive perception systems. To address the corresponding wireless data offloading issues, employing 5G networks presents a potentially viable solution\cite{yuan2022network}. Although Wi-Fi offers a more seamless and well-supported user experience, 5G promises a strategic trade-off by delivering reliable performance and resilience under high client density, making it a more suitable solution for manufacturing environments. 5G networks claim substantially higher achievable bandwidth compared to previous generations, facilitating swift and efficient transmission of large data utilized by CV. Additionally, 5G architectures are designed to minimize latency, a crucial point during real-time object detection and dynamic robot trajectory adjustments~\cite{urbaniak2024distributed}. However, despite the advantages of wireless solutions, 5G technologies have not yet reached full industry-grade maturity. Correspondingly, 5G devices capable of handling high-bandwidth CV tasks, such as industrial sensors or sensor platforms, remain commercially unavailable.
Current industrial implementations still heavily rely on external 5G edge gateways or commercial off-the-shelf (COTS) network adapters, which reintroduce cabling and mobility constraints~\cite{5gacia_machinevision_2025}.

Robotic grasping is a widely researched topic in adaptive robotics, mainly utilizing computer vision models to identify object parts, estimate their positions and orientations, and compute feasible grasp points. Ainetter and Fraundorfer~\cite{ainetter_end_end_2021} propose a novel end-to-end CNN architecture that jointly achieves high-quality grasp detection for parallel-plate grippers and semantic segmentation. Gou et al. introduce RGBD-Grasp, which decouples 7-DoF grasp detection into RGB-based orientation prediction and analytic depth-based grasp parameter estimation~\cite{gou_rgb_2021}. Horváth et al.~\cite{horvath_mogpe_2023} propose a real-time two-stage multi-object grasp pose estimation method that factorizes grasp prediction into object detection and class-specific orientation estimation. 
Low-latency computer vision models like YOLO11\footnote{Newer versions of the YOLO family, including YOLO12~\cite{tian2025yolo12} and YOLO26~\cite{ultralytics_YOLO26}, have been introduced after the completion of this work.}~\cite{ultralytics_YOLO11} also support oriented bounding boxes (OBB)---along with the traditional axis-aligned bounding boxes (AABB)---which can be effectively used for orientation estimation~\cite{cong2025improving}. The YOLO family adopts a unified, single-stage detection architecture~\cite{redmonYouOnlyLook2016}, enabling high inference speed while maintaining competitive detection accuracy. These properties make YOLO-based detectors well suited for real-time robotic perception tasks.

Deep learning-based models such as YOLO require large annotated datasets, whose collection is often costly and impractical. Transfer learning mitigates this limitation through approaches such as (i) fine-tuning pretrained models on the target domain and (ii) leveraging sim-to-real synthetic data in zero- or few-shot learning settings to reduce dependence on real-world annotations~\cite{horvath_object_2023}.

Related to hand-aware computer vision models, recent works \cite{leonardi2024syntheticdatausefulegocentric},~\cite{10936005} showed that synthetic data can significantly improve hand and hand–object interaction detection, including approaches based on fully synthetic datasets and diffusion-driven augmentation. These methods typically rely on complex generative pipelines, while lightweight compositing strategies, such as placing segmented real hands onto diverse backgrounds to approximate deployment conditions remain unexplored.

To the best of our knowledge, existing literature lacks a comprehensive approach that simultaneously addresses these infrastructure and perception-related challenges as a fully integrated system. As a result, there remains a gap regarding how fully wireless, portable workcells can be effectively deployed in HRC environments while supporting human safety. Conceptually, existing adaptive HRC frameworks rely on external 5G edge-gateways or COTS network adapters, which reintroduce cabling constraints. Our approach completely solves this issue by removing the need for external wires routed along the arm. By integrating a multi-sensor platform directly onto the cobot flange, this work shifts the networking bottleneck to the wireless edge. Furthermore, at the perception level, we propose a lightweight domain-randomization and compositing strategy instead of the complex generative pipelines in the current hand-aware models. This conceptual shift enables a high-speed YOLO-based vision pipeline that simultaneously provides oriented grasp-pose estimation and robust human hand detection for the latency and computational constraints of a barrier-free workcell.

\section{SOLUTION APPROACH} \label{sec:solution} 
To address the high variability and dynamic nature of remanufacturing workflows, we propose a highly adaptable HRC system. The proposed solution focuses primarily on the part manipulation element of remanufacturing workflows, however, it is easily extendable to other processes by using different manual tools and end effectors. 

\begin{figure}[thpb]
      \centering
      \includegraphics[width=\linewidth]{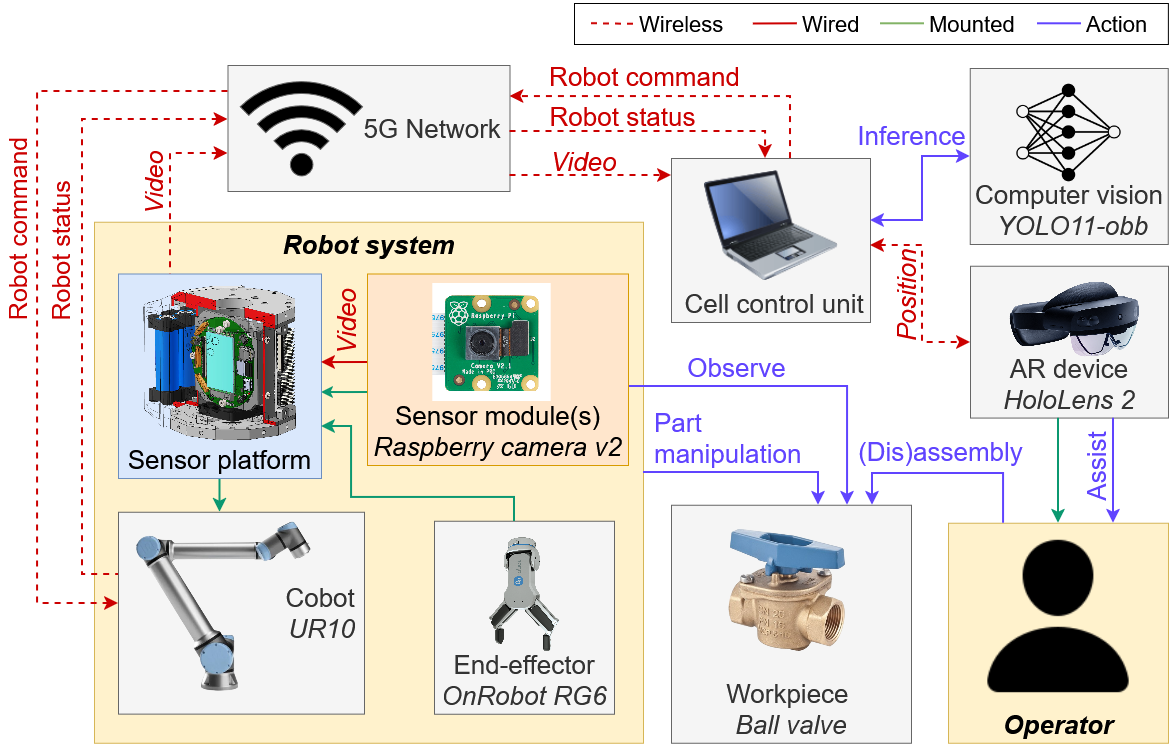}
      \caption{Generic and use case-specific components of the proposed 5G-based HRC system architecture. Specific components part of the case study are highlighted in italics.}
      \label{fig:sys_architecture}
    \end{figure}

The architecture (see Fig.~\ref{fig:sys_architecture}) features a cobot equipped with a 5G-based, battery-powered multi-sensor platform to ensure architecture modularity. The wireless arrangement eliminates power and data cabling, which maximizes robot mobility and enables rapid spatial reconfiguration for various layouts and remanufacturing workflows. The multi-sensor platform captures sensor data (such as video streams) and transmits it over a low-latency 5G network to support CV tasks, including object and hand detection. Leveraging this, the system empowers dynamic robot control, hand-aware safety features, and AR assisted operator guidance.

\subsection{Multi-Sensor Platform Design}
To enable a wireless and reconfigurable HRC cell, we engineered a 5G-based multi-sensor platform, incorporating a custom adapter to mount directly to standard ISO 9409\mbox{-}1 flange~\cite{iso9409_1_2004} of cobots. The design prioritizes system modularity, which allows the use of standard wired sensors and enables rapid changes of sensors and cameras, without the cumbersome cable routing and arrangement along the robot arm. In addition, establishing thermal stability is also of key importance. 

\begin{figure}[thpb]
      \centering
      
      \includegraphics[width=\linewidth]{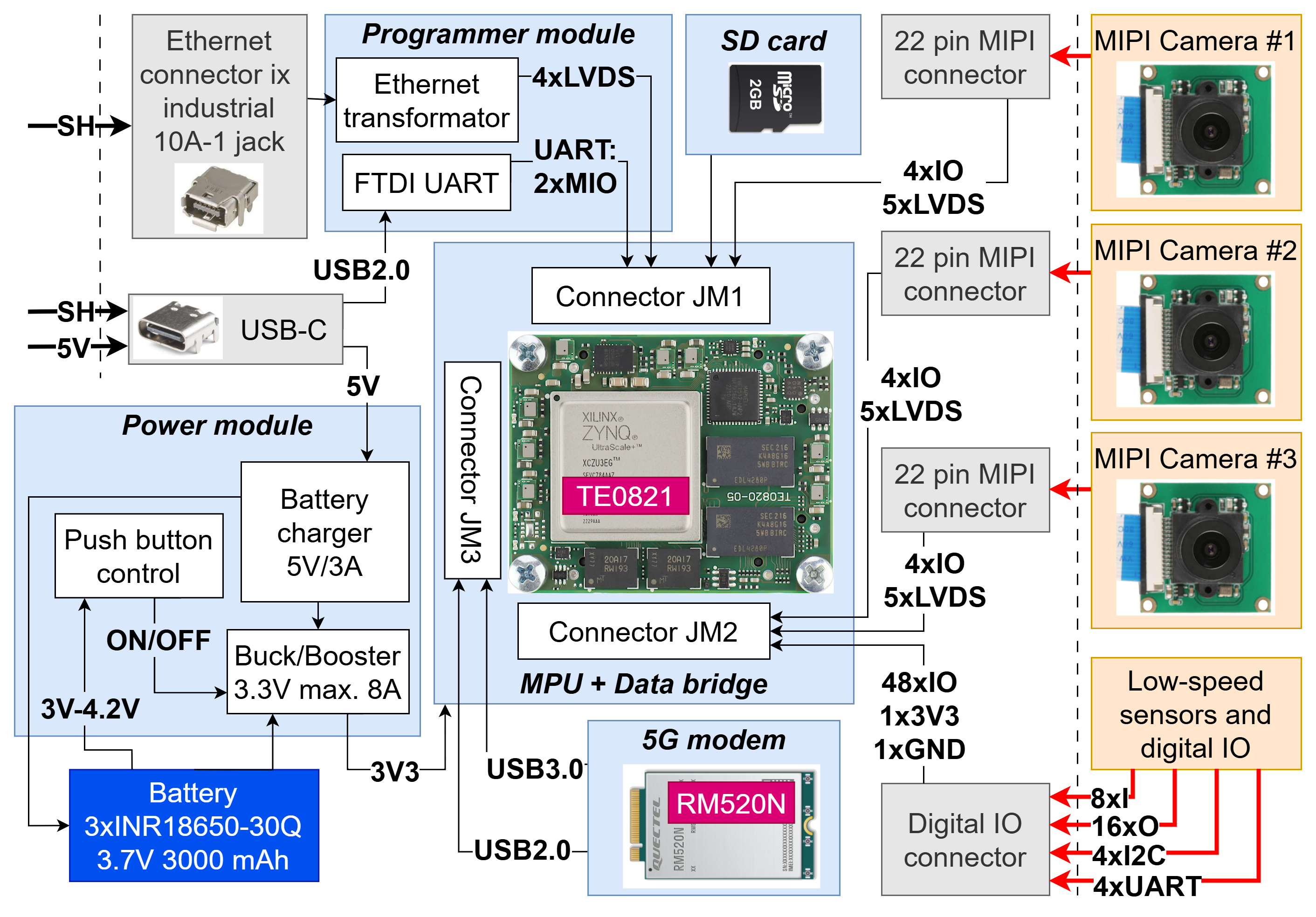}
      \caption{Structure of the 5G-based multi-sensor platform.}
      \label{fig:sp_architecture}
   \end{figure}

Our sensor platform structure (see Fig.~\ref{fig:sp_architecture}) integrates COTS components onto a custom-designed carrier board. We selected both the Trenz Electronics TE0821 System-on-Module (SoM) and the Quectel RM520N 5G modem for their high-performance technical profiles, specifically leveraging the TE0821 SoM’s integrated hardware video encoder. Unlike software-based alternatives, this hardware acceleration minimizes bandwidth and latency in wireless edge environments, allowing the system to process and transmit three independent camera streams without computational overload.

This custom carrier board serves as the system’s motherboard, hosting the COTS modules and the power management logic designed for autonomous battery operation. Additionally, the board features a wide range of peripheral interfaces, including a 1G Ethernet port and a 50-pin expansion connector. Of these pins, 24 are used as configurable general-purpose IOs (GPIOs), configured in the prototype as 8 digital inputs and 16 digital outputs, 4 I2C and 4 UART, for low-speed sensor integration. Although this study primarily uses the platform as a wireless vision node, we included low-speed interfaces to support future hardware expansions. This design allows for the addition of force, pressure, or proximity sensors, enhancing both precision and robustness of assembly tasks while supporting close-range operator safety. To maintain the compact form factor complying with the load and size capacity of cobots, the electronics utilize the smallest available surface-mounted components. Finally, our software stack bridges low-level hardware control via hardware description language (HDL) with an application layer developed in Python and C for efficient data streaming and system monitoring.

The mechanical housing was optimized for metal 3D printing production. Thermal management proved to be the primary design constraint as excessive heat can accumulate around the central computing unit. To address this, we used a vertical electronic arrangement to facilitate efficient airflow, and integrated active cooling via a built-in fan. The 3D model of the multi-sensor platform is shown in Fig.~\ref{fig:sp_prototype} (right).

Furthermore, usability testing with early models motivated us to shift from internal power sources to externally mounted battery packs. This modular approach allows rapid power swapping to support extended operation hours and significantly reduces the thermal load within the housing. A dedicated software module monitors on-chip thermal sensors, providing warnings and safety shutdowns if operating temperatures exceed safe threshold. Fig.~\ref{fig:sp_prototype} (left) presents the final prototype mounted on a robot arm.

\subsection{Computer Vision}
\begin{figure}[htbp]
    \centering
    \begin{minipage}{0.48\linewidth}
        \centering
        \includegraphics[width=\linewidth]{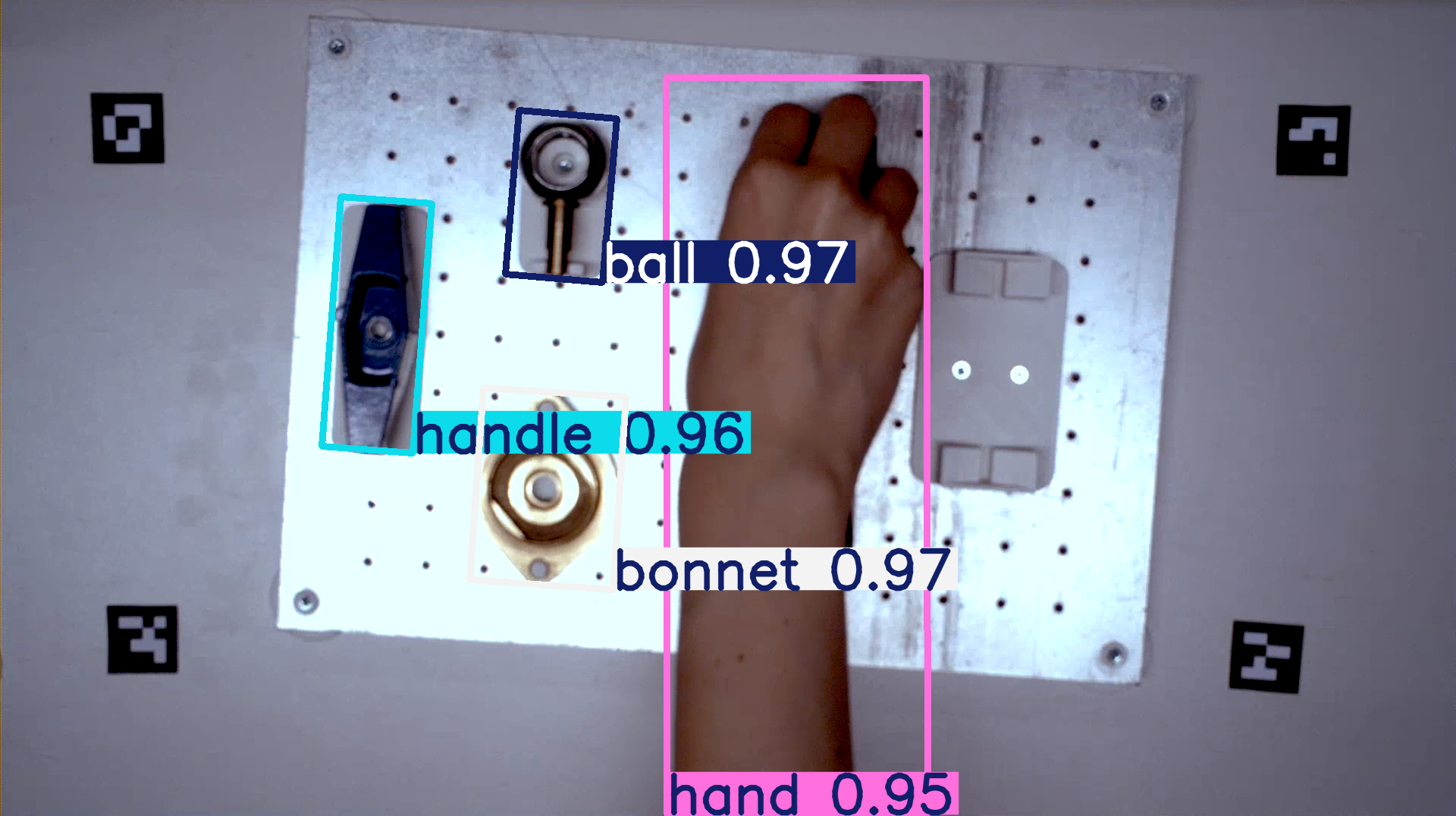}
    \end{minipage}
    \hfill
    \begin{minipage}{0.48\linewidth}
        \centering
        \includegraphics[width=\linewidth]{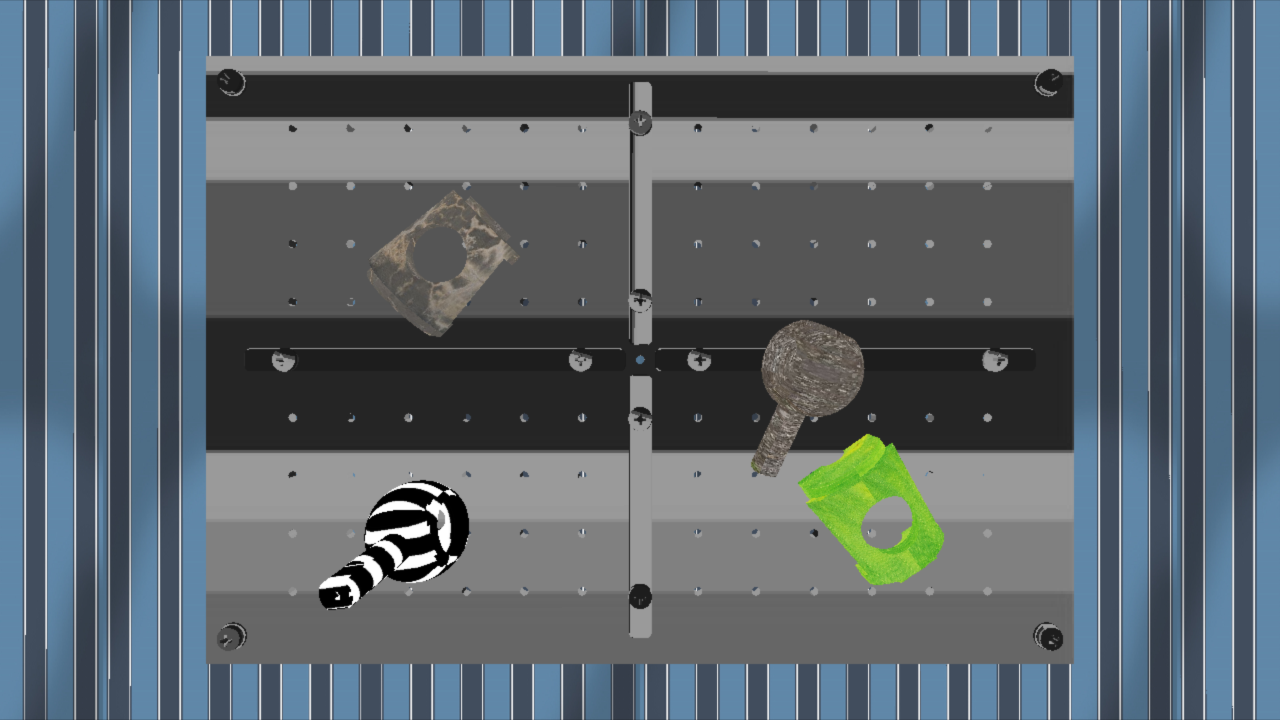}
    \end{minipage}
    \caption{Left: Object detection with highlighted OBBs and hand recognition results on a real image. Right: A domain-randomized synthetic image for preliminary training.}
    \label{fig:synthetic_real}
\end{figure}

In HRC scenarios, reliable detection of industrial components is essential to support both robotic manipulation and operator guidance. We formalize our CV problem as $Y  = f(X)$, where $X \in \mathbb{R}^{H \times W \times C}$, where $H,W \in \mathbb{N}$ are the width and height of the image and $C \in \mathbb{N}$ is the number of channels. The output $Y$ is described by $\boldsymbol{Y} = \left\{ \left(\boldsymbol{b}_i, c^{\text{class}}_i, p^{\text{con}}_i \right) 
\;\middle|\; i = 1,2,\ldots,N \right\}$

where $\boldsymbol{b}_i = [x_i,\, y_i,\, w_i,\, h_i,\, \theta_i]$ represents the 2D oriented bounding box (OBB), $c^\text{class}_i \in \mathbb{N}$ is the class label, and $p^\text{con}_i$ is the confidence score of the $i$\textsuperscript{th} detection, while $N \in \mathbb{N}$ is the number of detected objects. An example image with the detected objects is shown in Fig.~\ref{fig:synthetic_real} (left).

To solve this formalized problem under strict low-latency constraints, we adopted a YOLO-based architecture. As our component-wise choice, this detector was selected because of its favourable trade-off between detection accuracy, inference speed, and training simplicity. Compared with two-stage detectors, YOLO provides significantly lower latency, while avoiding the higher computational cost and training complexity often associated with transformer-based approaches, making it well suited for real-time oriented object detection. 

Accurate estimation of the object’s planar position and its orientation around the axis normal to the worktable surface is required to determine the appropriate target pose. In addition, hand-awareness is also essential to the CV pipeline: object bounding boxes are only considered when no human hand is detected in the scene, whereas the presence of a hand suppresses inference to prevent occlusion induced errors and unsafe target poses. 

Using artificial markers on the worktable with known positions in the robot base frame, image-space detections can be transformed into robot-space poses. At runtime, the markers are detected in the camera image together with the OBBs. A planar homography~\cite{kazemi2010visualservoing} is computed between the image and the worktable plane, allowing the detected object position and in-plane orientation to be mapped into the robot base frame. It is noted that a hand–eye calibration procedure could be used as an alternative to obtain the camera-to-robot transformation without relying on markers.

In order to reduce labelling burdens, we generated domain-randomized synthetic images following the approach proposed in~\cite{horvath_object_2023}. An example image is shown in Fig.~\ref{fig:synthetic_real} (right). These support the creation of an initial annotated dataset, enabling the training of a preliminary detection model. To further enhance robustness and improve generalization to varying backgrounds, lighting conditions, and other domain shifts, additional data augmentation techniques were applied, as well as hand-specific augmentation strategies.

\section{CASE STUDY AND RESULTS} \label{sec:results}
To validate the practical viability and robustness of our proposed solution, we deployed the system as a comprehensive experimental testbed across diverse geographical and network environments. We implemented the reconfigurable workcell across two international facilities at HUN-REN SZTAKI (Hungary) and the University of Oslo (Norway). By utilizing private SA, public SA, and private NSA 5G networks at these sites, we focused on evaluating the system's portability, network resilience, and adaptability to domain shifts. Furthermore, with this multi-site deployment, we intended to demonstrate the barrier-free cell's capacity to support a spectrum of future research and development scenarios, including remanufacturing, operator training, and HRC user studies.

\subsection{Experimental Setup}
In order to demonstrate the capabilities of our 5G-based reconfigurable cell, the proposed architecture was applied to a remanufacturing use case involving the disassembly and re-assembly of a ball valve. This product was selected because it mimics typical industrial complexities while remaining portable and it is intuitive enough to teach to operators, or to participants during future user studies. 

In this implementation, following the architecture outlined in Fig.~\ref{fig:sys_architecture} (with case study-specific components in italics), the part-manipulating cobot is a UR10 robotic arm equipped with the multi-sensor platform, onto which a Raspberry Pi v2 camera is mounted to monitor the shared workspace. While the sensor platform has a built-in modem, the control laptop and robot controller connect to the 5G network via a ZTE MC 888 external modem. During the remanufacturing workflow, the camera streams video data over the low-latency network to the cell control unit, which evaluates the scene to continuously detect ball valve parts and human hands. Fig.~\ref{fig:sp_prototype} (left) displays a completed cell.

The HRC-based remanufacturing workflow selected for our case study was set up as follows. First, during the disassembly phase, the human operator manually dismantles the ball valve and places the components randomly into the shared workspace, while the robot performs the physical sorting tasks. To achieve this, the CV module performs OBB detection to dynamically estimate precise grasping poses for the cobot. On the other hand, in the re-assembly phase, the operator manually constructs the ball valve. The system uses the CV output and a predefined assembly graph to identify the next required component, dynamically guiding the operator step-by-step via the AR assistant. When the operator finished the assembly, the CV output can provide the robot the target poses for feeding new parts for the operator.

Throughout both scenarios, hand detection acts as a safety mechanism---if the hands of the operator enter the shared workspace, the system triggers a safety protocol and stops the cobot or reduces its speed. A supplementary video demonstrating the reconfigurable workcell is available at https://youtu.be/zobin6oytGk.

\begin{figure}[htbp]
    \centering

        \centering
        \includegraphics[width=\linewidth]{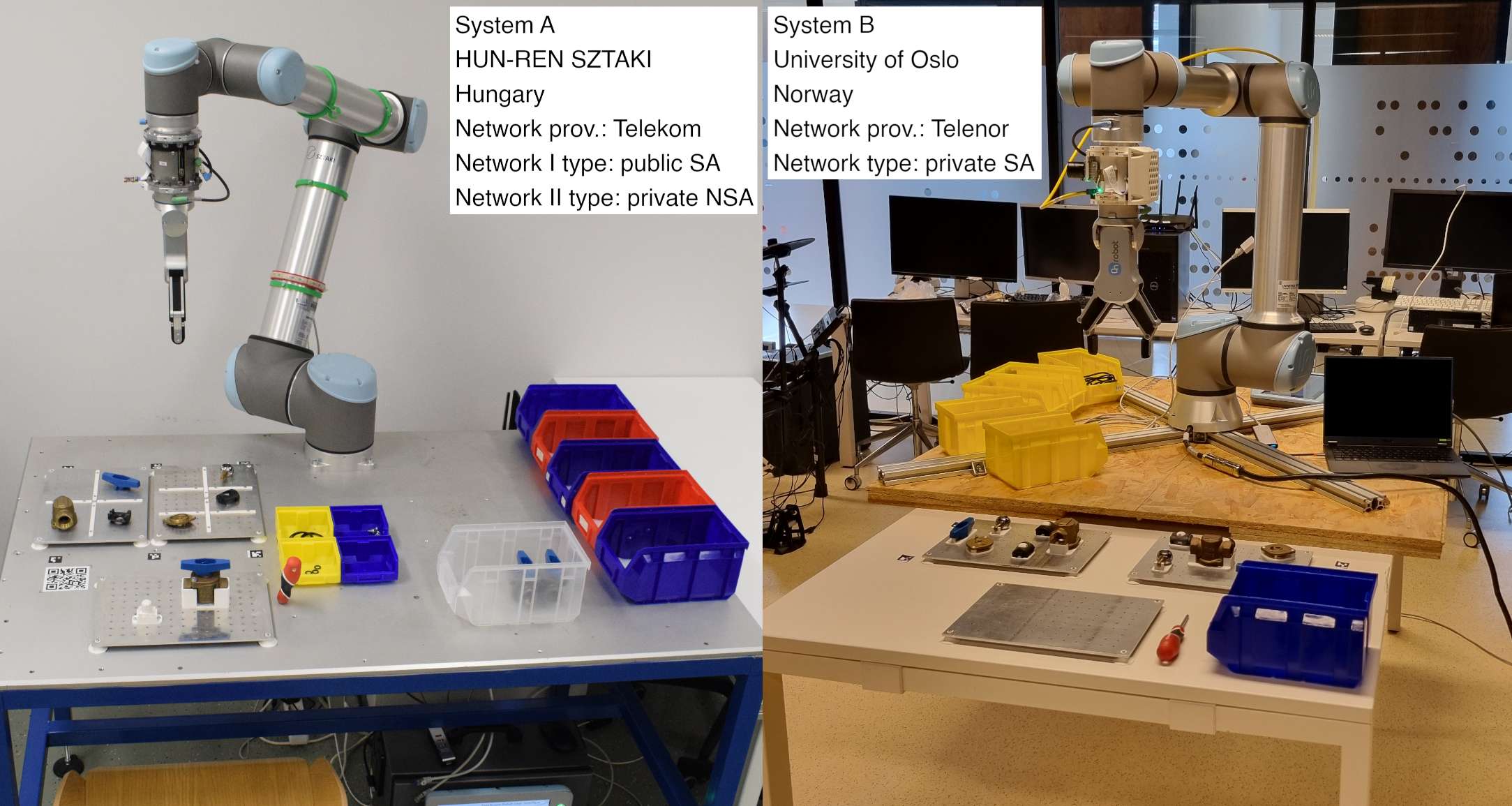}

    \caption{Finished robot cell setups for the case study, showing different lighting and background conditions. Left: System A at HUN-REN SZTAKI, Innovation and Demonstration Space (IDS), Hungary. Right: System B at University of Oslo, Sustainable Immersive Networking Lab (SINLAB), Norway.}
    \label{fig:cells}
\end{figure}

To validate the system's robustness and portability, we deployed our architecture at two separate geographical sites (see Fig.~\ref{fig:cells}) and conducted full experimental evaluations at both facilities (see Sections~\ref{sec:network_eval} and~\ref{sec:cv_eval}). By hosting the HRC workcells at different institutes across two countries, we were able to test the framework over heterogeneous set of 5G networks, while simultaneously challenging the CV model with diverse lighting and background conditions. 

To transition from the qualitative case study to the quantitative evaluation, the following sections provide a detailed component-wise analysis of our proposed system.

\subsection{5G Network Performance Analysis}
\label{sec:network_eval}

To analyse the impact of network architecture components, the presented system architecture (shown in Fig.~\ref{fig:sys_architecture}) was further extended with a test computer and a synthetic load generator for network performance testing.

The two implemented systems were operated under different network providers and configurations according to Fig.~\ref{fig:cells}. In System A, a primary and a secondary network of Telekom were deployed. The primary one (I) is a public, SA network, where connections are established via publicly routable addresses, utilizing the commercial network infrastructure. The secondary network (II) is a hybrid, NSA one, with a dedicated hotspot and direct communication between all devices. On the other hand, being in a different country, the network at System B was established by Telenor. This was set up as a private, yet SA network, resembling the combination of System A-I and A-II.

For the evaluation, we employed four key performance indicators (KPIs): (i) response time, measured as round-trip time of packets, (ii) jitter, measured as the difference between minimum and maximum response times, (iii) throughput, and (iv) maximum transmission unit size (MTU).
Three experimental scenarios were defined to incorporate all key components of the system, measuring in both directions.
All experiments were executed using fully automated shell scripts, utilizing \textit{ping} and \textit{iPerf}. Experimental results are summarized in Table~\ref{tab:kpi_results}.

Although network providers and device specifications claimed compatibility, the setup and testing of both systems revealed network--device pairing issues. In all setups, the external 5G modem connected to the cobot and the cell controller operated perfectly, confirming the 5G network signals. Yet, the built-in 5G modem of the multi-sensor platform was not able to pair well with networks A-I and B. In case of System A-I, connection was established and it was stable, but measured response times were extremely high. In case of System B, the multi-sensor platform was only able to connect to the network, but the connection could not be maintained long enough to perform the experiments, as it was regularly dropped from the network. 

In both scenarios, despite extensive troubleshooting in coordination with network providers, configuration attempts failed to yield measurable improvements in the built-in modem connectivity. This can be attributed to a hardware–network configuration mismatch, which forced us into performing experiments less representative of the sensor platform: for System A, meaningful results coming only from the secondary network (A-II); for System B, connecting the sensor platform with the same external 5G modem used for the cell control unit and the robot.

In the first experimental scenario (Case \#1), connection tests were executed between the sensor platform and test computer and the robot and test computer. The results show that unloaded communication works well in all scenarios (including System A-I) with standard packet sizes of 1420~B.

In the second experimental scenario (Case \#2), baseline performance measurements were performed using the same hardware combinations as in Case \#1, with a set of single-thread and multi-thread bandwidth values (10, 100, 200 Mbps). Robot-side measurements show consistently good results, achieving low response times and jitters, while continuously matching the bandwidth setting in throughput.
Although the network itself was not overloaded, in higher bandwidth cases (100 and 200 Mbps), the sensor platform hardware turned out to be the bottleneck, achieving only 12 Mbps throughput and up to 105 ms response time in case of System A-II and 108 Mbps and up to 19 ms in case of System B. As discussed previously, in case of System A-I, the network--sensor platform pairing produced extremely poor results (0.5 Mbps throughput and response times in the range of seconds).

Finally, in the third scenario (Case \#3), performance measurements were repeated while the load generator introduced an additional 50 Mbps network traffic. Robot-side results remained consistent with Case \#2, showing a slight reduction in the performance. 
Regarding the sensor platform, results indicate that simultaneous loads did moderately reduce the performance for the 100 and 200 Mbps bandwidth cases: 9 Mbps throughput and up to 131 ms response time in case of System A-II, as well as 67 and 81 Mbps and up to 14 ms in case of System B, respectively.

The achieved results show that standard and high definition video streams (typically requiring throughput in a range of 1-5 Mbps per camera) are suitable for the current capabilities of the multi-sensor platform. However, the reached results did only meet 5G-related expectations under specific network--device pairings, highlighting unforeseen interoperability challenges that limit the potential present-day applicability of the 5G technology. Consequently, compatibility issues and corresponding network and device specifications need to be carefully analysed and validated to achieve a truly industry-grade 5G system.

\setlength{\tabcolsep}{3.5pt}
\begin{table}[h]
\caption{Required network KPIs and corresponding results: Mean values of maximum transmission unit size (MTU), response time (RT), jitter (JIT) and throughput (TP).}
\label{tab:kpi_results}
\begin{center}
\begin{tabular}{|c|l|l|c|c|c|c|c|c|}
\hline
\rotatebox[origin=lB]{90}{\makecell[l]{Case \#}} & \multicolumn{1}{c|}{\rotatebox[origin=lB]{90}{\makecell[l]{Bandwidth\\{}setting [Mbps]}}} & \multicolumn{1}{c|}{\rotatebox[origin=lB]{90}{\makecell[l]{KPI}}} & \rotatebox[origin=lB]{90}{\makecell[l]{Required}} & \rotatebox[origin=lB]{90}{\makecell[l]{System B\\Sensor platform\,}} & \rotatebox[origin=lB]{90}{\makecell[l]{System B\\Robot}} & \rotatebox[origin=lB]{90}{\makecell[l]{System A-I\\Sensor platform}} & \rotatebox[origin=lB]{90}{\makecell[l]{System A-II\\Sensor platform}} & \rotatebox[origin=lB]{90}{\makecell[l]{System A-I\\Robot}}\\
\hline
\multirow{2}{*}{1} & \multirow{2}{*}{---} & MTU[B] & 1420$\leq$ & 1420 & 1420 & 1420 & 1420 & 1420\\
& & RT[ms] & 30$\geq$ & 27 & 27 & 20 & 24 & 15\\
\hline
\multirow{9}{*}{2} & \multirow{3}{*}{10} & RT[ms] & 30$\geq$ & 19 & 25 & 2187 & 12 & 27\\
& & JIT[ms] & 20$\geq$ & 6 & 8 & 973 & 6 & 16\\
& & TP[Mbps] & 20$\leq$ & 10 & 10 & 0.5 & 10 & 10 \\
\cline{2-9} 
& \multirow{3}{*}{100} & RT[ms] & 30$\geq$ & 15 & 23 & 2326 & 104 & 43\\
& & JIT[ms] & 20$\geq$ & 6 & 7 & 1083 & 55 & 21\\
& & TP[Mbps] & 20$\leq$ & 99 & 100 & 0.5 & 12 & 100 \\
\cline{2-9} 
& \multirow{3}{*}{200} & RT[ms] & 30$\geq$ & 15 & 24 & 1324 & 105 & 52\\
& & JIT[ms] & 20$\geq$ & 5 & 7 & 2025 & 54 & 26\\
& & TP[Mbps] & 20$\leq$ & 108 & 200 & 0.5 & 12 & 166 \\
\hline
\multirow{9}{*}{3} & \multirow{3}{*}{10} & RT[ms] & 30$\geq$ & 19 & 27 & 2573 & 108 & 25\\
& & JIT[ms] & 20$\geq$ & 7 & 8 & 1675 & 67 & 14\\
& & TP[Mbps] & 20$\leq$ & 10 & 10 & 0.5 & 9 & 10\\
\cline{2-9} 
& \multirow{3}{*}{100} & RT[ms] & 30$\geq$ & 14 & 22 & 2192 & 131 & 40\\
& & JIT[ms] & 20$\geq$ & 4 & 8 & 1190 & 77 & 21\\
& & TP[Mbps] & 20$\leq$ & 67 & 100 & 0.6 & 9 & 100\\
\cline{2-9} 
& \multirow{3}{*}{200} & RT[ms] & 30$\geq$ & 14 & 25 & 2202 & 121 & 48\\
& & JIT[ms] & 20$\geq$ & 5 & 7 & 1159 & 65 & 23\\
& & TP[Mbps] & 20$\leq$ & 81 & 198 & 0.6 & 9 & 163\\
\hline
\end{tabular}
\end{center}
\end{table}

\subsection{Computer Vision Implementation and Evaluation}
\label{sec:cv_eval}
To construct the training dataset for the CV module, the proposed semi-automatic labelling strategy was applied to reduce labelling burdens. First, the official YOLO11-obb checkpoint was fine-tuned on a combination of 4000 synthetically generated images and 10 manually annotated real images. An example generated image is shown in Fig.~\ref{fig:synthetic_real} (right). The synthetic data generation process was tailored to oriented object detection following the methodology described in~\cite{horvath_object_2023}. A preliminary model was trained on the synthetic data, and then was used to pre-annotate an additional 440 real images. These pre-annotations required only minor manual corrections to produce the final labelled dataset. Notably, the synthetic images were used solely for training this preliminary model and were not included in the training of the final model.

\begin{figure}[htbp]
    \centering
    \includegraphics[width=\linewidth]{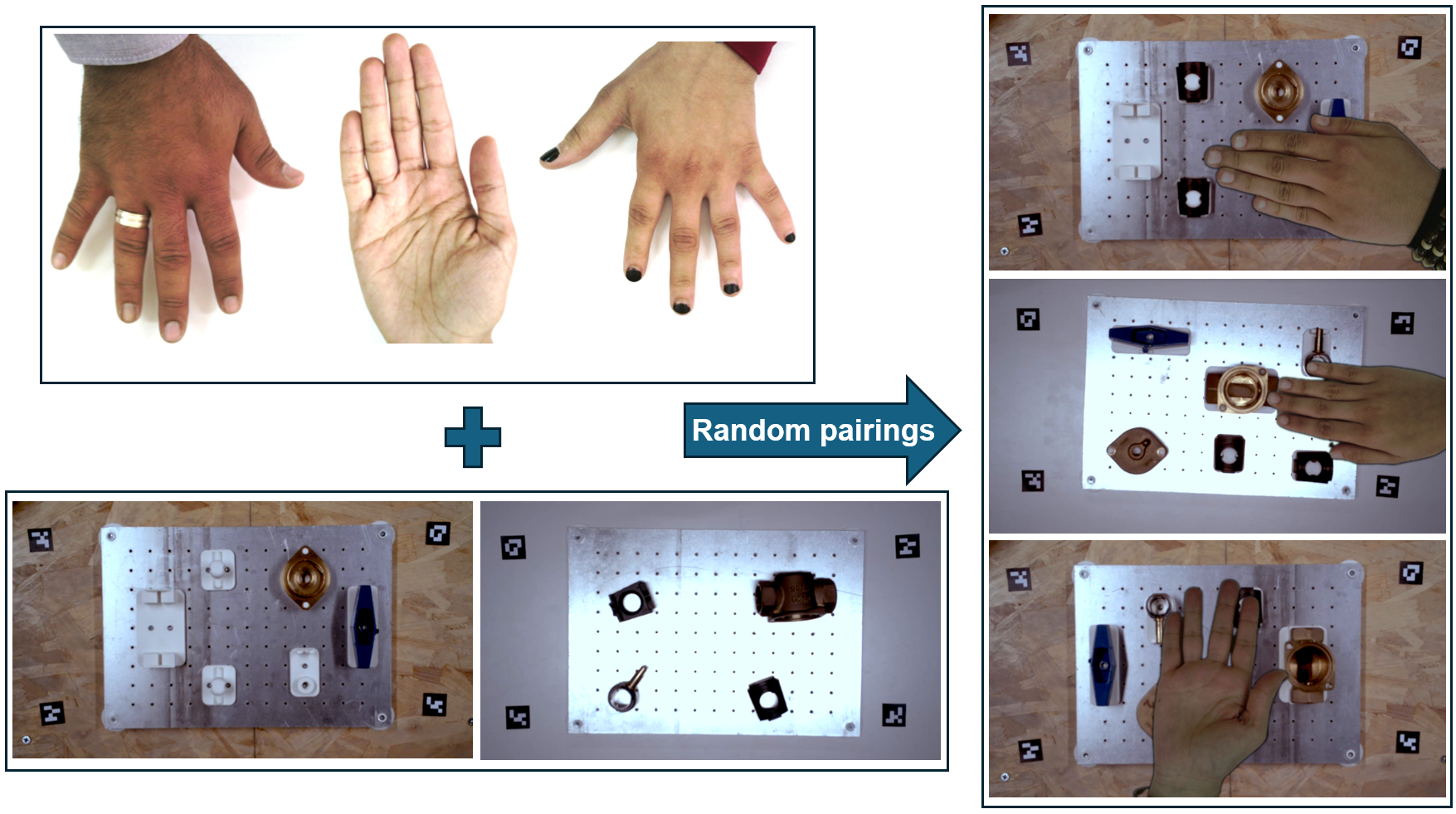}
    \caption{Illustration of the hand augmentation pipeline. Hand images are segmented and randomly composited onto different laboratory backgrounds captured under varying lighting conditions.}
    \label{fig:hand_augmentation}
\end{figure}

To enable hand recognition, hand detection was incorporated into the baseline model as an additional class. An initial dataset was set up by capturing 198 images using the equipped camera, depicting one or two hands from multiple viewpoints, including palmar and dorsal sides. In addition, to ensure robustness, an augmentation pipeline was introduced as follows. From the publicly available 11k Hands dataset~\cite{afifi201911kHands}, 1000 images were randomly selected, providing variation in gender, age, skin tone, and other accessories such as rings, watches and nail polish. To better approximate the experimental setup, the uniform white background of these images was removed via segmentation, and the hands were composited onto 107 background images captured under varying lighting conditions, within the laboratory setting. Hand–background pairings were randomized, and additional transformations were applied, including stochastic scaling, rotation, Gaussian blur, and hue–saturation–value perturbations. An example of the composite image is shown in Fig.~\ref{fig:hand_augmentation}.

Object detection performance was measured using the \mbox{mAP@50--95} metric. To evaluate training stability, the model was trained six times with different random seeds, and results are reported as the mean $\pm$ standard deviation. The final model achieved an overall \mbox{mAP@50--95} of 97.74 $\pm$ 0.10\%, demonstrating both highly accurate performance and low variance across runs. 

Although implemented as an additional class within the YOLO11-obb model, hand detection was evaluated as a binary presence classification task. The test dataset comprised 246 images, of which 195 contained visible hands. Using the same six training runs described above, precision, recall, and F1-score were computed at a fixed confidence threshold of 0.5. Precision remained 1.000 (std = 0.000) across all runs, while recall achieved a mean of 0.9991 (std = 0.0021). A single false negative was observed in one of the six runs, whereas all other evaluations produced no misclassifications. The resulting mean F1-score was 0.9996 (std = 0.0010), indicating highly reliable hand presence detection.

Runtime was measured on the NVIDIA RTX 4060 Laptop GPU installed in the cell control unit. The first inference exhibited a startup overhead of up to 100 ms, however in steady-state runtime measurements the model achieved a mean inference time of 12.5 ms with a standard deviation of 1.7 ms, suitable for safe HRC tasks.

\subsection{System Limitations}
While the proposed architecture successfully demonstrates the viability of a wireless, adaptive HRC workcell, some practical limitations must be acknowledged:

\begin{itemize}
    \item Network interoperability: During our cross-border evaluation, the built-in 5G modem of the multi-sensor platform failed to maintain stable connections with both public (A-I) and private (B) SA networks due to hardware-network configuration mismatches. This issue reflects the ongoing interoperability challenges of current 5G deployments.
    
    \item Occlusion: Because the system maps coordinates using artificial markers, extensive occlusion from an operator's arm blocking the view can temporarily disrupt pose estimation and affect grasping accuracy.
    
    \item Battery constraints: Although the multi-sensor platform features rapid battery swapping, the current prototype's battery capacity presents a practical limitation. In our user study experiment case, the system sustained about 80-120 mins of powered-on time, with the camera actively running for workspace monitoring and detection 50\% of the time.
\end{itemize}

\section{CONCLUSION} \label{sec:conclusion}
This work presents a modular, wireless, reconfigurable HRC workcell specifically designed to overcome the limitations imposed by restrictive cabling and the dynamic nature of remanufacturing environments. By developing an innovative, battery-powered, 5G-based multi-sensor platform and offloading computationally intensive perception tasks to the edge, we effectively eliminated physical cables, while maintaining low latency suitable for safe and adaptive human-machine interaction. Acting as a comprehensive experimental testbed, the proposed system removes infrastructure barriers, and supports a spectrum of research and development scenarios, including remanufacturing, assembly and disassembly, operator training, HRC research, and user studies.

Within this research, we also provided a robust computer vision pipeline designed to generalize across domain shifts, even between international facilities. Leveraging a YOLO11-obb architecture augmented by a synthetic data generation strategy, the system achieved reliable detection of oriented grasping poses and human hands in different lighting conditions and backgrounds (with \mbox{mAP@50--95} of 97.74 $\pm$ 0.10\% and a mean inference time of 12.5 ms). This low-latency perception layer directly facilitates a responsive augmented reality (AR) assistant, closing the loop between the robot and the operator through dynamic AR guidance.

We validated our architecture through cross-border experiments across heterogeneous 5G network environments, specifically testing private Standalone (SA), public SA, and private Non-Standalone (NSA) deployments. Our results demonstrate the system's portability and low-latency capabilities, achieving average round-trip response times down to 12 ms in case of compatible network--device pairings, while simultaneously highlighting that practical network resilience remains an open issue for real-world deployments. The performed experiments provide empirical evidence that with adequate combinations and correct configurations, modern 5G infrastructures can meet the strict timing requirements of current industrial systems. However, it also became apparent that ensuring compatibility can be an unpredictable, non-trivial task that can severely limit applicability. In order to become a reliable option for industrial application, 5G technology still needs to grow in maturity and improve in interoperability.

Future work will focus on expanding the hardware versatility and the application of our proposed architecture. We plan to develop a next-generation prototype of the multi-sensor platform that (i) integrates Wi-Fi connectivity alongside 5G to maximize deployment flexibility across different industrial infrastructures, (ii) incorporates an alternative built-in modem to resolve the observed network compatibility issues, and (iii) provides extended support for USB-based sensors. Additionally, because the multi-sensor platform acts as the mechanical intermediary between the cobot's flange and the end-effector, future iterations will include a comprehensive Finite Element Method stress analysis for structural evaluation under dynamic payloads and emergency stops. Finally, we aim to conduct extensive HRC user studies to evaluate operator cognitive load, safety perceptions, and overall system usability during remanufacturing workflows.

\addtolength{\textheight}{-12cm}   




\section*{ACKNOWLEDGMENT} 

The research was supported by EU HORIZON-JU-SNS-2022 Research and Innovation Programme under Grant Agreement No. 101096452 (IMAGINE-B5G) and Horizon Europe Research and Innovation project rEUman under Grant Agreement No. 101138930.
We would like to thank for the usage of HUN-REN Cloud (https://science-cloud.hu/) that significantly helped us achieving the results published in this paper.


\addtolength{\textheight}{20\baselineskip}

\bibliographystyle{IEEEtran}
\bibliography{IEEEabrv,refs.bib}

\end{document}